\documentclass{article}

\pdfoutput=1
\usepackage{cite}
\usepackage{arxiv}
\usepackage[utf8]{inputenc} 
\usepackage[T1]{fontenc}    
\usepackage{hyperref}       
\usepackage{url}            
\usepackage{booktabs}       
\usepackage{amsfonts}       
\usepackage{nicefrac}       
\usepackage{microtype}      
\usepackage{lipsum}
\RequirePackage{amsmath,amsfonts,amssymb}
\RequirePackage{graphicx,xcolor}

\title{Ensemble Clustering for Graphs}

\author{
  Valérie Poulin \\
  Tutte Institute for Mathematics and Computing \\
  Ottawa, Canada \\
  \texttt{vpoulin@gmail.com} \\
   \And
  François Théberge \\
  Tutte Institute for Mathematics and Computing \\
  Ottawa, Canada \\
  \texttt{theberge@ieee.org} \\
}

\begin{document}
\maketitle

\begin{abstract}
We propose an ensemble clustering algorithm for graphs (ECG), which is based on the Louvain algorithm and the concept of consensus clustering. We validate our approach by replicating a recently published study comparing graph clustering algorithms over artificial networks, showing that ECG outperforms the leading algorithms from that study. We also illustrate how the ensemble obtained with ECG can be used 
to quantify the presence of community structure in the graph.
\end{abstract}


\section{Introduction}
\label{sec:1}
Many data-sets are relational in nature, describing interactions between entities, such as
friendship networks, communications or geographical co-locations.
Most networks that arise in nature exhibit complex structure \cite{Girvan:2002,Newman:2003} with subsets of vertices densely interconnected
relative to the rest of the network, which we call communities or clusters.
Binary relational data-sets are typically represented as graphs $G=(V,E)$, where vertices $v \in V$
represent the entities, and edges $e \in E$ represent the relations between pairs of entities. For analyzing and exploring complex relational data-sets, graph clustering is commonly used.

In this paper, we propose ECG (Ensemble Clustering for Graphs), a graph clustering method based on the concept of co-association consensus clustering. We show that this approach
identifies very high quality clusters by replicating the study in \cite{Yang:2016} and comparing ECG against the best performing algorithms. We also demonstrate that ECG is stable despite the fact of being a randomize algorithm and that it reduces significantly the resolution limit problem, yielding a number of clusters very close to the ground truth partition size. Finally, ECG provides information about the strength of the associations between entities which can be used to determine the presence or absence of communities in the network.

\section{Related Work}
\label{sec:2}
Graph clustering aims at finding a partition of the vertices $V=C_1 \cup ... \cup C_l$ into good clusters.
This is an ill-posed problem \cite{Fortunato:2016}, as there is no universal definition of good clusters, leading to a wide variety of graph clustering 
algorithms \cite{Girvan:2002,Clauset:2004,Pons:2005,Newman:2006,Raghavan:2007,Reichardt:2006,Rosvall:2007,Blondel:2008},
with different objective functions.
Over the last decade, several studies were conducted to compare graph clustering algorithms \cite{Yang:2016,Fortunato:2016,Orman:2009,Lancichinetti:study:2009,Orman:2012}.
In a recent study \cite{Yang:2016}, several state-of-the art algorithms implemented in the {\tt igraph} \cite{Csardi:2006} package were compared over a wide
range of artificial networks generated via the LFR benchmark \cite{Lancichinetti:2008}.

Those studies indicate that the multilevel Louvain (ML) algorithm \cite{Blondel:2008} offers one of the best tradeoff between the quality of the clusters it produces and its speed.
There are however some issues with this algorithm:  it is unstable, i.e.,  
results from successive runs on the same data can vary a lot, and it suffers from the well-known resolution limit problem \cite{Fortunato:2007}, i.e., the ML algorithm often yields a coarsening of the true structural clusters. Methods such as re-weighting edges with respect to presence in short cycles \cite{Berry:2009} do help to
improve the stability, but are computationally expensive.

Ensemble clustering, often referred to as consensus clustering, offers a wide selection of algorithms \cite{Vega-Pons:2011,Li:2010} for multivariate data. The general idea behind consensus clustering is to combine several partitions
over the same data-set into a final partition.
This strategy has been motivated by
the success of ensemble methods for supervised learning \cite{Freund:1997, Breiman:2001} which, despite their simplicity, are often among the best classification methods. Consensus clustering consists of two steps: (i) the {\it generation}
step, where several partitions of the data is obtained, and (ii) the {\it integration} step, where the final partition is computed using a consensus function \cite{Vega-Pons:2011}. 
The consensus function is typically based on co-occurrences of objects in clusters \cite{Strehl:2002,Fred:2005} (co-association consensus) or on finding a median partition \cite{Li:2010} which maximizes the similarity
with all partitions obtained in the generation step.

\section{Algorithm Description}
\label{sec:3}
Let $G=(V,E)$ be a graph where $V=\{1,2,...,n\}$ is the set of vertices, and 
$E \subseteq \{(u,v) ~|~ u, v \in V,~ u < v\}$ is the set of edges. 
We consider undirected graphs.
Edges can have weights $w(e)>0$ for each $e \in E$. 
For un-weighted graphs, we let $w(e)=1 ~ \forall e \in E$.
The 2-core of a graph $G$ is its maximal subgraph whose vertices have degree at least 2.
Let $P_i = \{C_i^1,...,C_i^{l_i}\}$ be a partition of $V$ of size $l_i$. We refer to the $C_i^j$
as {\it clusters} of vertices. We use $\mathbf{1}_{C_i^j}(v)$ to denote the indicator function for $v \in C_i^j$.

\subsection{Review of Multilevel Louvain}

The ML algorithm produces a hierarchy of partitions where the level-$i$ partition is a coarsening of the level-$(i-1)$ partition in the hierarchy. The level-0 partition is the partition consisting of all singletons on $V$. To obtain each subsequent level, the following two phases are repeated:
\begin{enumerate}
    \item Perform a random permutation of the vertices. For each vertex, evaluate the change in modularity obtained by removing this vertex from its current community, and placing it in the community of its neighbor which yields the highest increase in modularity. Perform community change according to the largest gain in modularity for this vertex (if any) otherwise leave the vertex in its current community. This is repeated until no gain of modularity is achieved, which may require the vertices to be visited more than once.
    \item Build a weighted quotient graph using the partition obtained in the first phase.
\end{enumerate}
Those two phases are repeated until no further significant improvement in modularity can be achieved.
This algorithm has several advantages: it is very efficient, it returns a hierarchy of partitions, and the number of clusters need not be specified beforehand. However, randomization of the vertex ordering
tends to yield very different solutions albeit with similar modularity, so the algorithm is unstable.
This statement is true for the final level partition of ML, but it applies even more to the lower levels of the hierarchy. Obtaining several randomized level-$1$ partitions on a given graph yields many weakly correlated (fine) partitions of the graph. This is what we use in the generation process of ECG.

\subsection{ECG Algorithm}

The ECG algorithm is a consensus clustering algorithm for graphs. As previously mentioned, its generation step consists of independently obtaining $k$ randomized level-$1$ ML partitions: ${\mathcal P} = \{P_1,...,P_k\}$. Its integration step is performed by running ML on a weighted version of the initial graph $G=(V,E)$. The weights are obtained through co-association, i.e., the weight of an edge $e=(u,v) \in E$ is defined as 

$$
W_{\mathcal P}(u,v) = \left\{
\begin{array}{cc}
w_* + (1-w_*) \cdot \left(\frac{\sum_{i=1}^k v_{{P}_i}(u,v)}{k}\right), & \mbox{if} (u,v) \mbox{ is in the 2-core of } G \\
w_*, & \mbox{otherwise}
\end{array}
\right.
$$

\noindent where $0 < w_* < 1$ is the minimum ECG weight and $v_{P_i}(u,v) = \sum_{j=1}^{l_i} \mathbf{1}_{C_i^j}(u) \cdot \mathbf{1}_{C_i^j}(v)$ indicates if the vertices $u$ and $v$ co-occur in the same cluster of $P_i$ or not. Note that all $W_{\mathcal P}(e) \in [w_*,1]$ for $e \in E$ and the minimum weight $w_*$ is assigned to edges whose endpoints are in different clusters for all partitions in $\mathcal{P}$, or to edges outside the 2-core. We discuss this choice in Section \ref{sec:5}. 
This process can be seen as an efficient alternative to re-weighting via the enumeration 
of short cycles such as triangles, since 
edges which belong to several short cycles will see their vertices often put in same level-1 ML clusters.
It has been shown that using {\it weak} clustering algorithms in the generation step can produce high quality 
results \cite{Topchy:2005}. 
Running a single level of the ML algorithm is a good example a weak learner,
where vertices are grouped into many small clusters. 
When running the ECG algorithm, the size $k$ of the ensemble and the minimum edge weight $w_*$
are the only parameters that need to be supplied.

\section{Comparison Study Revisited}
\label{sec:4}
In this section, we re-visit a recently published comparison study of graph clustering algorithms \cite{Yang:2016},
comparing the best performing algorithms from that study with ECG algorithm.
In that study, eight different state-of-the-art graph clustering algorithms are evaluated, all of which 
are included in the {\it igraph} package.
The algorithms are compared on graphs generated with the LFR benchmarks for 
undirected and unweighted graphs and with non-overlapping communities.
The key parameter when generating an LFR graph $G$ is the {\it mixing parameter} $\mu$, which sets the expected
proportion of edges in $G$ for which the two endpoints are in different communities. The parameter $\mu$ 
can be interpreted as the expected proportion of noisy edges in a pure community graph (graph with disconnected communities).

\subsection{Comparison measures}

An LFR graph $G$ is generated along with its ground-truth partition $P_G$. Hence, we can define the accuracy of a partition $P$ on $G$ via a  similarity measure $sim(P,P_G)$. 
Most measures used as graph partition similarities are in fact set partition similarity
scores either based on pair counting \cite{Hubert:1985,Albatineh:2011} or on Shannon's
information \cite{Meila:2007,Vinh:2009,Vinh:2010}.
It is highly recommended to use their adjusted versions \cite{Hubert:1985,Vinh:2009}.
In our study, we report the non-adjusted normalized mutual information (NMI) measure as
used in \cite{Yang:2016}, but also the 
adjusted rand index (ARI) recommended when 
the number of data points is large relative to
the number of parts in the partition \cite{Romano:2016}.
As demonstrated in a recent paper \cite{Poulin:2018}, it is also very important to include a {\it graph-aware} measure when comparing graph partitions. Those measures
are shown to be complementary to the set partition measures and superiority in both types of measures is shown to be necessary to assess the superiority of an algorithm over another.
As a consequence, we will also report the adjusted graph-aware rand index (AGRI) 
in our experiments.

\subsection{Main results}

Comparisons over several graphs are presented in \cite{Yang:2016}, with graph sizes (number of vertices)
ranging from $233$ to $31,948$.
For most experiments, the graph sizes are in the set $n \in \{233, 482, 1000,
3583, 8916, 22186\}$, and the mixing parameter is in the set $\mu \in \{0.03, 0.06, ..., 0.72, 0.75\}$.
For each combination of parameters $(n,\mu)$, we generate 100 LFR graphs with the parameters listed in Table 1 of \cite{Yang:2016}.
We use the {\it igraph} (version 0.7.1) implementation of the graph clustering algorithms, running in Python version 3.6. For ECG, we use our own python implementation \cite{Theberge:Code:2018}.
For the similarity measures (NMI, ARI), we use the implementation in
{\it scikit-learn} \cite{sklearn} (version 0.19.0) while we implement the AGRI
measure in Python as described here \cite{Poulin:Code:2018}. 
For NMI, several normalizations exist \cite{Vinh:2010}; we use the {\it sum of entropy}
as in \cite{Yang:2016}. Other normalization yield very similar plots and equivalent conclusions.

We are interested in algorithms that can handle larger graphs, not only graphs with a few hundred edges. From \cite{Yang:2016} and other studies \cite{Lancichinetti:study:2009,Orman:2012},
three algorithms stand out as good choices
for small as well as large graphs: ML, the Walktrap \cite{Pons:2005} (WT) algorithm, 
and the InfoMap (IM) algorithm \cite{Rosvall:2007}.
We compare ECG to these algorithms.

\begin{figure}[ht]
\vspace{1.5cm}
\centering
\includegraphics[width=14cm]{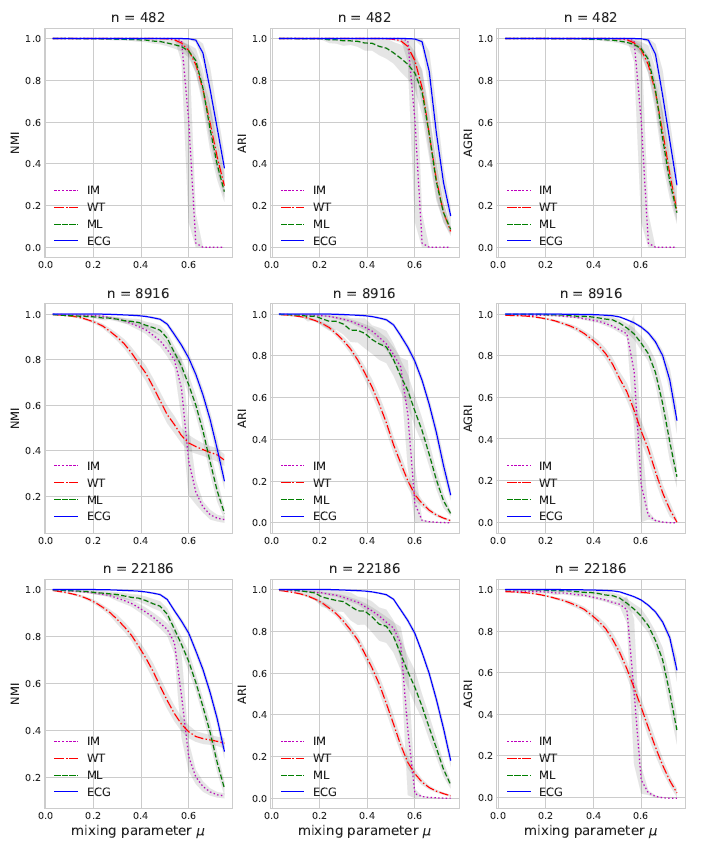}
\caption{
Accuracy results with respect to the ground truth communities for the 4 graph clustering algorithms considered for the LFR graphs where $n \in \{482, 8916, 22186\}$. 
Comparison is done via the NMI, ARI and AGRI similarity measures. 
Each curve shows the mean over 100 different LFR graphs for each value
of $\mu$, the shaded area is the standard deviation.
In all cases, we see that ECG outperforms all other algorithms.
}
\label{fig:all-scores}
\end{figure}

In Figure \ref{fig:all-scores}, we compare the accuracy of each algorithm over
three of the graph sizes $n$, and over the range of mixing parameter $\mu$. 
The conclusions are the same for the other choices of $n$.
From these plots, we see that WT and IM do well for small and mid-range values of $\mu$,
and ML is generally a better choice, in particular with noisier graphs.
The ECG algorithm outperforms all other algorithms, and the difference is more pronounced
for larger graphs.

\begin{figure}[ht]
\vspace{1cm}
\centering
\includegraphics[width=14cm]{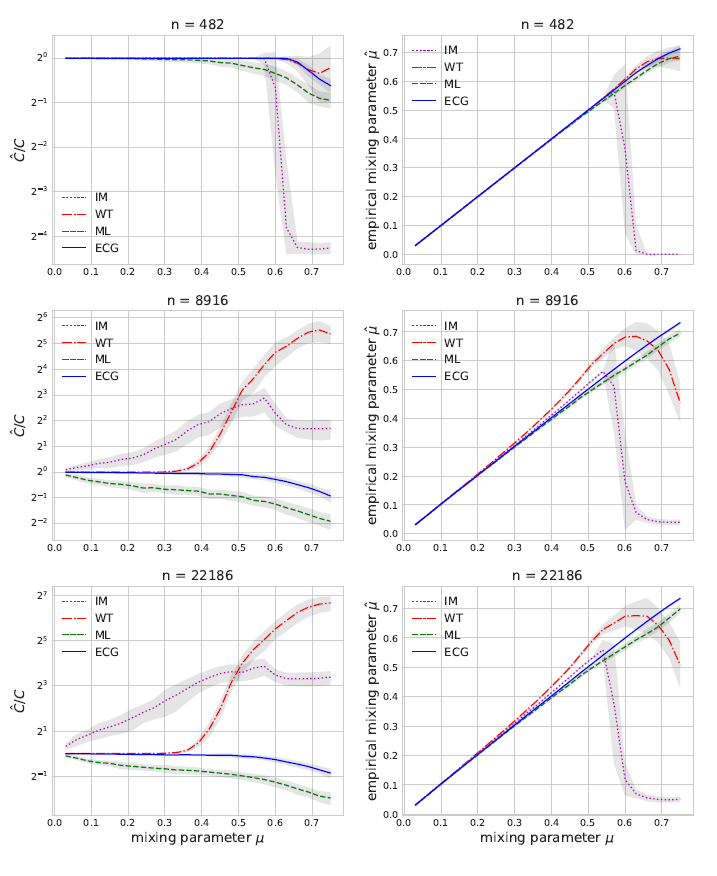}
\caption{
Comparing the 4 graph clustering algorithms on the LFR graphs with graph sizes $n \in \{482, 8916, 22186\}$.
In the left column, we look at the ratio of the number of computed clusters vs true communities, $\hat{C}/C$ for each value of $\mu$.
We see that ECG remains very close to the desired value $\hat{C}/C=1$. 
In the right column, we compute the empirical mixing parameter ${\hat{\mu}}$ for each value
of $\mu$. We see that the ECG algorithm stays very close to the curve ${\hat{\mu}} = \mu$.
Each curve shows the mean over 100 different LFR graphs for each value
of $\mu$, the shaded area is the standard deviation.
}
\label{fig:all-etc}
\end{figure}

For each algorithm, we compute the ratio of the number of
clusters it found $\hat{C}$, divided by the true number of communities $C$ from the LFR generator.
We plot the results in Figure \ref{fig:all-etc}.
This is computed for the same set of values $n$ and $\mu$ as in Figure \ref{fig:all-scores}.
We also compare the empirical mixing parameter $\hat{\mu}$ computed from the clustering obtained with each algorithm.
From those plots, we see that ML has a
tendency to yield a smaller number of clusters, which would indicate a coarsening of the true communities.
WT, on the other hand, tends to return a larger number of clusters, 
likely indicating that true communities are broken up to produce a refinement of the true partition.
IM also yields a larger number of clusters for the larger graphs.
The ECG algorithm returns a number of clusters that is remarkably similar to the true value for a wide
range of the mixing parameter $\mu$. 

Looking at the empirical mixing parameter $\hat{\mu}$, we see that IM is close to the true parameter $\mu$ only for less noisy graphs while ML is generally quite good, with ECG producing slightly
better results, in particular for larger graphs.

\clearpage

\clearpage

\clearpage

\section{Discussion}
\label{sec:5}
In the previous section, we saw the clear advantage of the ECG algorithm over a wide
range of tests in terms of accuracy, number of communities found, and empirical mixing parameter.
In this section, we clarify some reasons for this success by verifying that known issues of ML are indeed solved with ECG. Moreover, we analyze the parameter selection robustness and, finally, we make use of the information contained in ECG weights as a way of determining if networks do contain strong community structure or not.

\subsection{Resolution limit and stability issue}

\begin{figure}[h]
\centering
\includegraphics[width=15cm]{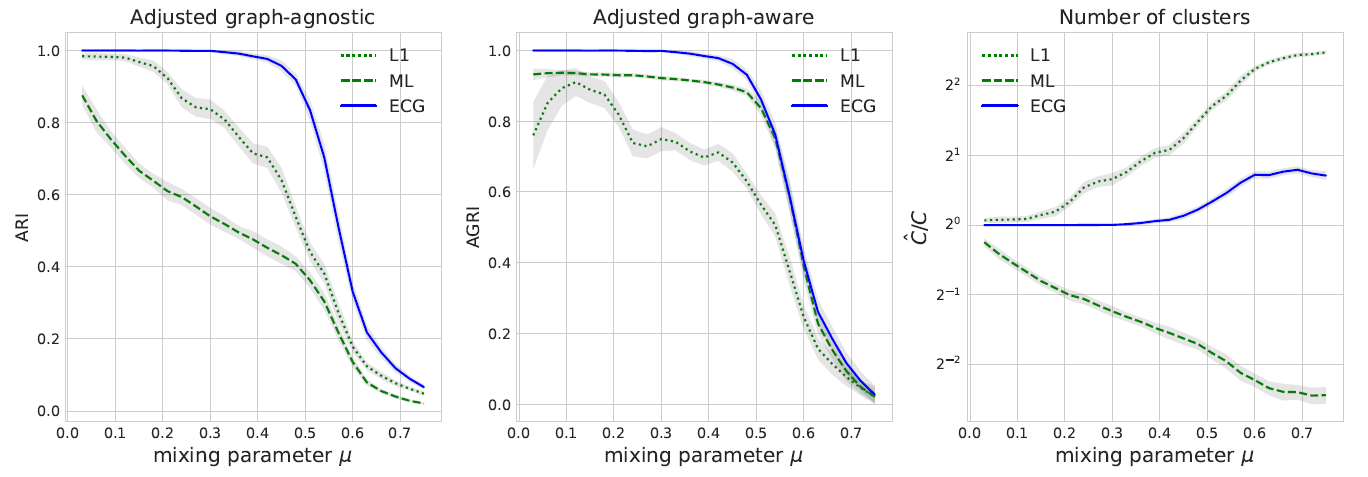}
\caption{For each $\mu$, 
we generated 50 LFR graphs of size 1000 with vertices of degree 8 and
community size in the range [10,15].
With respect to the two measures ARI and AGRI, ECG is more accurate than both ML 
L1, while conclusions between ML and L1 is different depending on the measure.
This can be explained by looking at the third plot, where we see that ML tends to return a smaller number of clusters
than the true number, while L1 returns a larger number. On the other hand, ECG is quite close to the true number.
}
\label{fig:louvain}
\end{figure}

Common issues with graph clustering algorithms are that (i) clusters can be broken up,
leading to a refinement of the true partition, or (ii) clusters can be amalgamated, leading to
a coarsening of the true partition. 
It was shown \cite{Poulin:2018} that graph-agnostic measures such as the 
ARI will give high scores for refinements of the true partition, while graph-aware
measures such as the AGRI will give high scores for a coarsening of the true partition.
ML algorithm tends to produce a coarsening
of the true partition. On the other hand, if we stop ML algorithm at the first level,
which we denote as L1, a refinement of the true partition into several smaller clusters is obtained.
Looking at the example in Figure \ref{fig:louvain}, 
we see that if we only use ARI, we conclude that L1 is superior to ML,
while if we only use AGRI, we conclude the opposite.
One of the main advantage of ECG is that the size of the partition it returns is much closer
to the true size, thus avoiding breaking up or merging clusters. 
As a consequence, ECG outperforms ML and L1
with respect to both measures.

Another advantage of ECG is the fact that the results 
tend to be more stable than with other randomized algorithms such as ML.
This is illustrated in the first plot of Figure \ref{fig:parameters}, where we use the LFR graphs with
$n=482$ vertices from the previous section. For each graph, we ran both ECG and ML twice,
and we compute the ARI score between the two partitions we obtained. We plot the mean ARI for
the different values of the mixing parameter.
The stability of ML starts to drop around $\mu=0.3$ and above, while ECG remains very stable until
about $\mu=0.6$.

\subsection{Parameter selection}

\begin{figure}[ht]
\centering
\includegraphics[width=15cm]{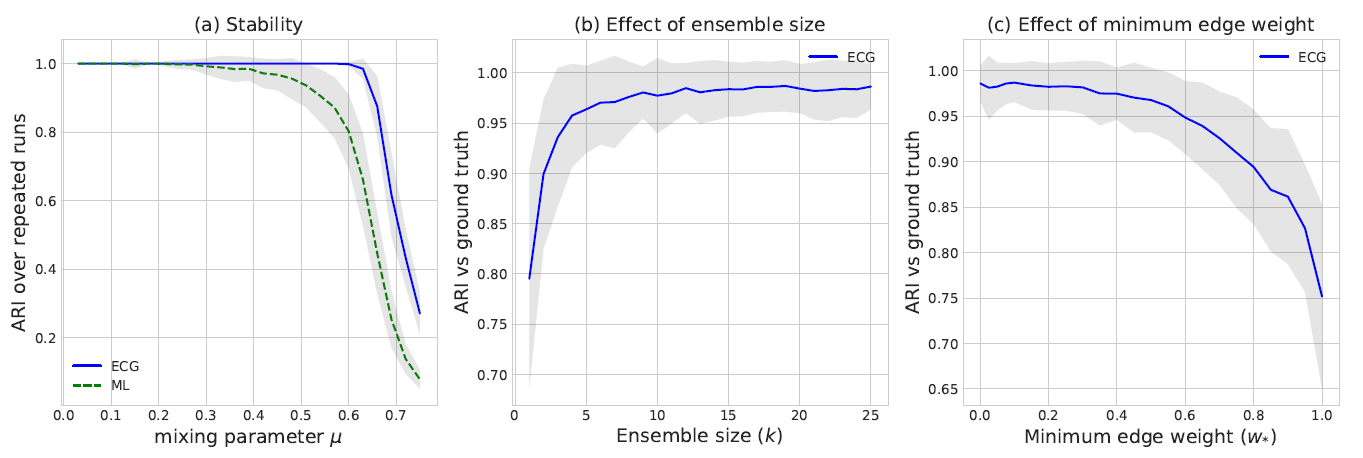}
\caption{In (a), we compare the stability of the ECG and ML algorithms by computing
the ARI measure on two runs over the same graph. We consider LFR graphs of
size $n=482$ from the previous section. We see that ECG remains stable over
a wider range of values for $\mu$. In (b) and (c), we look at the effect of the two
parameters in ECG: the ensemble size $k$ and the minimum weight $w_*$. We consider the
100 LFR graphs of size $n=482$ and $\mu=0.63$. 
We see that the default parameters ($k=16,~w_*=0.05$) are suitable, and that the algorithm is robust over the choice
of parameters. The standard deviations are indicated by the shaded areas.}
\label{fig:parameters}
\end{figure}

There are two parameters in the ECG algorithm. However, the results are not too sensitive
to their choice, and the default values are usually suitable.
In the second plot of Figure \ref{fig:parameters}, we illustrate the impact of the ensemble size $k$
over the LFR graphs of size $n=482$, with $\mu=0.63$. 
We see that as long as the ensemble is not too small (roughly $k \ge 8$),
the results are comparable. Larger ensemble size does not hurt in terms of accuracy, but could have an 
impact on the running time. In our experiments, we found the default value $k=16$ to be suitable.
In the third plot of Figure \ref{fig:parameters}, we look at the impact of the minimum weight parameter $w_*$.
Recall that a smaller value for $w_*$ gives more importance to the ensemble, and at the limit, setting
$w_*=1$ amounts to ignoring the ensemble altogether. From that plot, we see that a small value for $w_*$
is preferable. Note that we avoid setting $w_*=0$ which could modify the topology of the graph, as edges
with zero weight are dropped.  In our experiments, we found the default value $w_* = 0.05$ to be suitable.

\subsection{Detecting community presence}

With ECG, new edge weights are derived, which are used for the final ML run.
Recall that the weight given to an edge is proportional to the number of members in the ensemble
of partitions for which both vertices ended up in the same cluster. We make an exception for edges outside the
2-core, which are assigned the minimal weight $w_*$.
The rationale for this exception is that we aim at finding communities supported by shared relations 
and in that sense, trees (free cycle graphs) have no such communities. The non-2-core subgraph is in fact a forest of
trees (known as tendrils) each having a root in the 2-core. Without the above exception, these edges
would get high weights due to the fact that there is very little diversity in their predicted communities. 

\begin{figure}[ht]
\centering
\includegraphics[width=12cm]{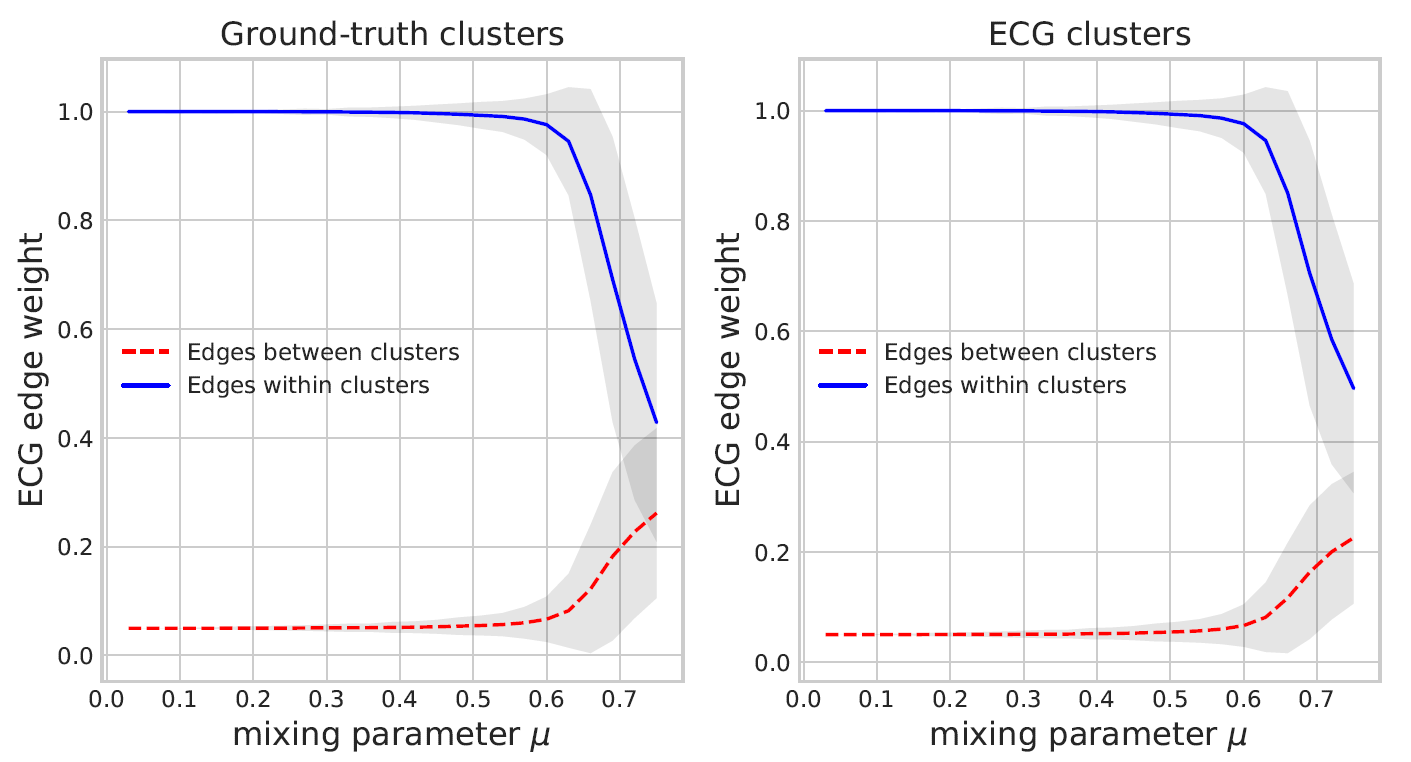}
\caption{We plot the distribution of edge weights respectively
for edges within or between clusters. We consider LFR graphs of
size $n=482$ from the previous section. On the left plot, the ground-truth LFR
clusters are used while on the right plot, we use the clusters returned by ECG.}
\label{fig:weights}
\end{figure}

With this in mind, we argue that weights generated by ECG can also be used to quantify 
the strength of clusters in the graph.
As an example, in Figure \ref{fig:weights}, we plot the distribution of ECG weights
for edges {\it within} clusters and edges {\it between} clusters. 
We consider both the ground-truth clusters, and the clusters found by ECG. 
This can be used as a diagnostic tool. For example,
for the graphs with $\mu \le .5$, the weights are very polarized (1 for edges within clusters, 0 
for the other edges), which indicates a strong community structure.
On the other hand, for large values of $\mu$, the weight distribution are not as well
separated, indicating a weak community structure.

\section{Conclusion}
\label{sec:6}
We proposed ECG, a new graph clustering algorithm based on the fast multilevel Louvain algorithm,
and on the concept of consensus clustering. We ran numerous experiments on artificial graphs, following a recently published study. All results obtained show an impressive accuracy and stability for ECG when compared to state-of-the-art algorithms. Moreover, ECG does not suffer from resolution error issue as much as other algorithms. It also provides weights which can be used to quantify the strength of the communities. Code is openly available \cite{Theberge:Code:2018}.


\bibliographystyle{unsrt}
\bibliography{references}

\end{document}